\documentclass[10pt]{article} % For LaTeX2e
\usepackage[preprint]{tmlr}

\usepackage{amsmath,amsfonts,bm}

\def\eqref#1{equation~\ref{#1}}
\def\1{\bm{1}}

\DeclareMathAlphabet{\mathsfit}{\encodingdefault}{\sfdefault}{m}{sl}
\SetMathAlphabet{\mathsfit}{bold}{\encodingdefault}{\sfdefault}{bx}{n}

\newcommand{\R}{\mathbb{R}}

\usepackage{microtype}
\usepackage{hyperref}
\usepackage{url}
\usepackage{booktabs}
\usepackage{amsmath}
\usepackage{amssymb}
\usepackage{mathtools}
\usepackage{algorithm}
\usepackage{algorithmic}
\usepackage{graphicx}
\usepackage{xcolor}
\usepackage{multirow}
\usepackage{array}
\usepackage{subcaption}
\usepackage{pgfplots}
\usepackage{cleveref}
\crefname{section}{Section}{Sections}
\Crefname{section}{Section}{Sections}
\crefname{figure}{Figure}{Figures}
\Crefname{figure}{Figure}{Figures}
\crefname{table}{Table}{Tables}
\Crefname{table}{Table}{Tables}
\crefname{equation}{Equation}{Equations}
\Crefname{equation}{Equation}{Equations}
\crefname{appendix}{Appendix}{Appendices}
\Crefname{appendix}{Appendix}{Appendices}

\pgfplotsset{compat=1.18}

\def\month{MM}  % Insert correct month for camera-ready version
\def\year{YYYY} % Insert correct year for camera-ready version
\def\openreview{\url{https://openreview.net/forum?id=XXXX}} % Insert correct link to OpenReview for camera-ready version

\newcommand{\lm}{LM}
\newcommand{\lambdainit}{\lambda_0}
\newcommand{\lambdainc}{\mu^+}
\newcommand{\lambdadec}{\mu^-}
\newcommand{\vect}[1]{\mathbf{#1}}

\title{Factored Levenberg--Marquardt for Diffeomorphic Image Registration: An efficient optimizer for FireANTs}

\author{ %
\name{Rohit Jena} \email rjena@seas.upenn.edu \\
\addr Department of Computer and Information Science\\
University of Pennsylvania\\
Philadelphia, PA 19104, USA
\AND
\name Pratik Chaudhari \email pratikac@seas.upenn.edu \\
\addr Department of Electrical and Systems Engineering\\
University of Pennsylvania\\
Philadelphia, PA 19104, USA
\AND
\name James C. Gee \email gee@upenn.edu \\
\addr Department of Radiology \& Penn Image Computing and Science Laboratory (PICSL)\\
Perelman School of Medicine, University of Pennsylvania\\
Philadelphia, PA 19104, USA
}

\begin{document}

\maketitle

\begin{abstract}
FireANTs~\citep{jena2025fireants} introduced a novel Eulerian descent method for plug-and-play behavior with arbitrary optimizers adapted for diffeomorphic image registration as a test-time optimization problem, with a GPU-accelerated implementation. 
FireANTs uses Adam as its default optimizer for fast and more robust optimization.
% We ask whether a second-order method can do better.
% The core challenge is that the Gauss--Newton Hessian for a 3D warp field with
% $10^6$--$10^7$ parameters is completely intractable to form or invert.
% We resolve this by factoring the Hessian approximation \emph{per voxel}:
% a rank-1 outer product of the local gradient yields a closed-form damped update
% that scales each voxel's step by the inverse of its local curvature.
However, Adam requires storing state variables (i.e. momentum and squared-momentum estimates), each of which can consume significant memory, prohibiting its use for significantly large images.
In this work, we propose a \textit{modified} Levenberg--Marquardt (LM) optimizer that requires only a single scalar damping parameter as optimizer state, that is adaptively tuned using a trust region approach.
The resulting optimizer reduces memory by up to $24.6\%$ for large volumes, and retaining performance across all four datasets.
A single hyperparameter configuration tuned on brain MRI transfers without modification to lung CT and cross-modal abdominal registration, matching or outperforming Adam on three of four benchmarks.
We also perform ablations on the effectiveness of using Metropolis-Hastings style rejection step to prevent updates that worsen the loss function.
\end{abstract}

% ============================================================
\section{Introduction}
\label{sec:intro}
% ============================================================

FireANTs~\citep{jena2025fireants} frames diffeomorphic image registration
as GPU-accelerated test-time optimization: a displacement field
$\vect{u} : \Omega \to \R^d$ is optimized by minimizing
\begin{equation}
  \mathcal{L}(\vect{u}) = \mathcal{L}_{\text{sim}}(F, M \circ \phi) +
  \mathcal{R}(\vect{u}),
\end{equation}
where $\mathcal{L}_{\text{sim}}$ is an image similarity loss and $\mathcal{R}$
regularizes the deformation $\phi = \mathrm{Id} + \vect{u}$.
Adam~\citep{kingma2014adam} is the default optimizer, and it works well.
But Adam is a quasi-second-order method: it exploits gradient \emph{history} through
exponential moving averages.
These moving averages can consume significant additional memory and runtime proportional to the voxels in the image, limiting their use for datasets with large volumes.
Although Adam is much cheaper than other second-order methods like L-BFGS~\citep{liu1989limited}, it could still introduce a significant overhead for large images.
This motivates the use of a more efficient second-order method that is not as memory-intensive.

Levenberg--Marquardt~(\lm)~\citep{levenberg1944method,marquardt1963algorithm}
is a well-known algorithm: a damped Gauss-Newton step that interpolates
between Newton's method (fast near optima) and gradient descent (robust far
from them).
However, the full $\vect{J}^\top\vect{J}$ Hessian is intractable for dense warp fields due to its scale: a $192^3$ displacement field has $\sim 5 \times 10^{13}$ parameters.
We show that the local structure of the registration loss makes a \emph{per-voxel}
rank-1 Hessian factorization both principled and sufficient, yielding a
closed-form update that is as cheap to compute as Adam yet demonstrates
second-order behavior.

To evaluate our approach, we conduct experiments on four publicly available benchmarks: brain MRI (LUMIR and OASIS), lung CT (NLST), and cross-modal abdominal MR-to-CT registration (Abdomen-L2R). We compare our factored Levenberg--Marquardt (\lm) optimizer against Adam, the default optimizer in FireANTs, using the same registration framework and similarity metric across datasets, showing slight improvements in performance without recording any additional state.
Our ablation on synthetically generated volumes demonstrate that \lm\ not only reduces memory consumption by up to $24.6\%$ for large volumes, but also matches or surpasses Adam in registration accuracy using a single hyperparameter configuration transferred across imaging modalities and organs.
We additionally provide ablations that highlight the effectiveness of a Metropolis-Hastings-style rejection step, which prevents loss-increasing updates and contributes to the optimizer's stability. 

% ============================================================
\section{Factored Levenberg--Marquardt for Warp Fields}
\label{sec:methods}
% ============================================================

\subsection{Compositive Diffeomorphic Updates with Adaptive Optimization}
\label{sec:diffeomorphic}

FireANTs maintains diffeomorphism throughout optimization via a compositive
update rule. 
Given a velocity $\vect{v}^{(t)}$ produced by the optimizer, the warp is
updated as
\begin{equation}
  \vect{u}^{(t+1)} = \epsilon\vect{v}^{(t)} + \vect{u}^{(t)} \circ
  (\mathrm{Id} + \epsilon\vect{v}^{(t)}),
  \label{eq:compositive}
\end{equation}
where $\epsilon$ is a small positive constant to ensure that $\mathrm{Id} + \epsilon\vect{v}^{(t)}$ is diffeomorphic.
We defer the reader to \citet{jena2025fireants} for the choice of $\epsilon$ for maintaining diffeomorphic behavior.
The quantity $\vect{u}^{(t)} \circ (\mathrm{Id} + \vect{v}^{(t)})$ resamples the
current warp at locations displaced by $\vect{v}^{(t)}$.
Optimization proceeds in a coarse-to-fine pyramid; at each level the warp
inherits the solution from the previous level, consistent with recommended practice in literature on multi-resolution optimization \citep{gee1991evaluation,jena2025fireants,avants2008symmetric,jena2025deep}.
Since diffeomorphisms lie on a Riemannian manifold and admit a group structure, deploying adaptive optimization (that are generally defined for Euclidean spaces) of $\vect{v}^{(t)}$ is a non-trivial problem.
~\citet{jena2025fireants} also proposes a novel Eulerian descent method for plug-and-play behavior of arbitrary optimizers for diffeomorphic image registration bypassing the need for expensive parallel transport operations by leverging the group structure of diffeomorphisms.
% Our \lm\ optimizer replaces the first-order velocity $\vect{v}$ with a
% curvature-scaled step while keeping this compositive integration intact.

% \subsection{Per-Voxel Hessian Factorization}
\subsection{Factored Levenberg--Marquardt for Diffeomorphic Registration}
\label{sec:lm}

\paragraph{Classical \lm\ and the scale problem.}
For a least-squares residual $\vect{r}(\boldsymbol{\theta})$ with Jacobian
$\vect{J}$, the \lm\ step solves
\begin{equation}
  \bigl(\vect{J}^\top \vect{J} + \lambda \vect{I}\bigr) \Delta\boldsymbol{\theta}
  = -\vect{J}^\top \vect{r},
  \label{eq:lm-classical}
\end{equation}
with $\lambda \geq 0$ the damping parameter.
For $\lambda = 0$ this is the Gauss--Newton step; as $\lambda \to \infty$
the step shrinks toward $-\vect{g}/\lambda$, recovering gradient descent.
For image registration, we typically optimize a single scalar loss function (ignoring regularization terms) $\mathcal{L}(\vect{u})$ which does not necessarily admit a squared loss form required for \lm.
However, one can convert the problem into a least-squares form by rewriting commonly used loss functions in a squared loss form.
\begin{enumerate}
    \item \textbf{Mean Squared Error (MSE):} This is already in a squared loss form.
    \item \textbf{Local Normalized Cross Correlation (LNCC):} This can be rewritten as a squared loss form by minimizing $(1 - \text{LNCC})^2$ instead of minimizing $-\text{LNCC}$.
    \item \textbf{Mutual Information (MI):} This can be rewritten as a squared loss form by minimizing $(\log_2(B) - \text{MI})^2$ instead of minimizing $-\text{MI}$, where $B$ is the number of bins for the parzen windowing method. Note that $\log_2(B)$ upper-bounds the discrete mutual information loss, so the squared loss function will always maximize the mutual information as well.
\end{enumerate}
Therefore, we only have a single scalar residual ${r}(\vect{u})$ (abbreviated as $r$ for brevity) and the Jacobian $\vect{J}(\vect{u})$ is a vector of the same length as the displacement field (and is actually the gradient of the loss function with respect to the displacement field).
Forming $\vect{J}^\top\vect{J}$ for a 3D displacement field with more than $\sim 10^7$ parameters is intractable.

\paragraph{Rank-1 factorization per voxel.}
The key observation is that the registration gradient
$\vect{g} \in \R^{H \times W \times D \times d}$
at each voxel $\vect{x}$ depends primarily on \emph{local} image content.
We therefore approximate the Gauss--Newton Hessian \emph{independently per
voxel} by the rank-1 outer product
$\hat{\vect{H}}_{\vect{x}} = \vect{g}_{\vect{x}} \vect{g}_{\vect{x}}^\top$.
The damped inverse of this $3\times 3$ matrix has a closed form (proof in \cref{sec:proof}):
\begin{equation}
  \Delta \vect{u}_{\vect{x}} = -\frac{r\vect{g}_{\vect{x}}}
  {\|\vect{g}_{\vect{x}}\|^2 + \lambda}.
  \label{eq:tile1}
\end{equation}
This is the \emph{pointwise update} (denoted as \texttt{tile\_size}=1 in our implementation): it normalizes
each voxel's gradient step by its magnitude, preserving direction while
suppressing updates at voxels with small, unreliable gradients.
No matrix storage or inversion is required beyond a scalar norm computation.

For richer spatial context, one can pool rank-1 approximations over a window of $(2w+1)^d$ voxels (where $w$ is the radius of the window, and $d$ is the dimension of the displacement field) to obtain a $d\times d$ positive semi-definite
$\hat{\vect{H}}_{\text{tile}}$, then apply
\begin{equation}
  \Delta \vect{u}_{\vect{x}} = -\bigl(\hat{\vect{H}}_{\text{tile}} +
  \lambda \vect{I}\bigr)^{-1} r\vect{g}_{\vect{x}},
  \label{eq:tilen}
\end{equation}
with the $3\times 3$ inverse computed via \texttt{torch.linalg.inv}.
In practice, \texttt{tile\_size}=1 matches or exceeds larger tiles at lower
cost (Section~\ref{sec:results}).

\paragraph{Adaptive damping.}
We adapt $\lambda$ after each step by comparing successive losses:
\begin{equation}
  \lambda^{(t)} = \begin{cases}
    \lambdainc \cdot \lambda^{(t-1)} & \text{if } \mathcal{L}^{(t)} >
    \mathcal{L}^{(t-1)} \quad \text{(bad step)}, \\
    \lambdadec \cdot \lambda^{(t-1)} & \text{otherwise} \quad
    \text{(good step)},
  \end{cases}
  \label{eq:lambda-update}
\end{equation}
with $\lambdainc > 1$ and $0 < \lambdadec < 1$.
This is a simple heuristic for trust-region-based algorithms.
The \lm\ step $\Delta\vect{u}$ is then normalized by the same choice of $\epsilon$ as in \citet{jena2025fireants} and integrated via Eq.~\eqref{eq:compositive}.

\paragraph{{\lm} with Rejection Criterion.}
Since {\lm} is a trust-region-based algorithm, we also implement an {\lm} variant with a rejection criterion. 
\citet{ranganathan2004levenberg} proposed a rejection criterion based on a quadratic model of the loss function, which can be expensive for large displacement fields. 
We implement a simpler rejection criterion that computes the quantity
\begin{equation}
\varrho_k = \frac{\mathcal{L}^{(k)} - \mathcal{L}^{(k-1)}}{\|\mathcal{L}^{(k-1)} - \mathcal{L}^{(k-2)}\|}
  \label{eq:delta-loss}
\end{equation}
and rejects the step if $\varrho_k > 1$, increases $\lambda$ according to \cref{eq:lambda-update} and retries the step with the incremented $\lambda$.
Intuitively, if the loss decreases, then $\varrho_k$ is negative, and we always accept the step.
We reject the step only if the loss degrades by more than the previous step's improvement (or tolerable degradation from the previous step).
This heuristic requires only the previous two loss values to be stored, and is computationally inexpensive.

\subsection{Connections to the Demons Algorithm}

The Demons algorithm of \citet{thirion1998image} and its diffeomorphic
extension \citep{vercauteren2009diffeomorphic} can be recovered as a special
case of per-voxel LM applied to MSE.
For MSE loss, the per-voxel residual is $r_{\vect{x}} = f(\vect{x}) -
m(\vect{x} + \vect{u})$ and the Jacobian of this scalar residual with respect
to $\vect{u}_{\vect{x}}$ is $\vect{J}_{\vect{x}} = -\nabla
m(\vect{x}+\vect{u})^\top$.
The \emph{Gauss--Newton} Hessian at voxel $\vect{x}$ is therefore
$\hat{\vect{H}}^{\mathrm{GN}}_{\vect{x}} = \vect{J}_{\vect{x}}^\top
\vect{J}_{\vect{x}} = \vect{n}_{\vect{x}}\vect{n}_{\vect{x}}^\top$,
where $\vect{n}_{\vect{x}} = \nabla m(\vect{x}+\vect{u})$.
Applying the Sherman--Morrison formula to the damped system
$(\vect{n}_{\vect{x}}\vect{n}_{\vect{x}}^\top + \lambda_{\vect{x}}\vect{I})\Delta\vect{u}_{\vect{x}} = r_{\vect{x}}\vect{n}_{\vect{x}}$
yields exactly
\begin{equation}
  \Delta\vect{u}_{\vect{x}} = \frac{r_{\vect{x}}\,\vect{n}_{\vect{x}}}
  {\|\vect{n}_{\vect{x}}\|^2 + \lambda_{\vect{x}}},
  \label{eq:demons}
\end{equation}
which is the Demons ``active forces'' update with $\lambda_{\vect{x}} = \alpha^2
r^2$.
A key difference is that Demons uses an `automatic', spatially varying, and
residual-dependent damping $\lambda_{\vect{x}} = \alpha^2 r^2$
chosen to regularize the optical-flow equation, whereas our global $\lambda$
is adapted across iterations via the trust-region rule of
\cref{eq:lambda-update}.
Furthermore, the Demons algorithm uses a `per-pixel' residual term $r_{\vect{x}}$ instead of a global residual $r$.
Finally, while Demons is defined specifically for MSE, our formulation
generalizes to other similarity metrics through the squared-loss reparameterizations in
\cref{sec:lm}, enabling second-order acceleration for a broader class
of similarity metrics.
However, the resemblence of the analytical form of the Demons algorithm to our \lm\ formulation hints at theoretical connections between the two methods, which we leave for future work.

% ============================================================
\section{Experiments and Results}
\label{sec:results}
% ============================================================

We evaluate on four publicly available benchmarks using FireANTs with \texttt{GreedyRegistration} and FusedCC similarity, comparing \textbf{Adam} (baseline, $\beta_1=0.9$, $\beta_2=0.999$),
and \textbf{Levenberg-Marquardt} ($\lambdainit=0.006$, $\lambdainc=1.5$, $\lambdadec=0.975$).
Benchmarks span brain MRI (LUMIR~\citep{lumir}, 38 pairs;
OASIS~\citep{oasis}, 19 pairs), lung CT (NLST~\citep{nlst}, 10 pairs), and
cross-modal abdominal MR$\to$CT (Abdomen-L2R~\citep{hering2022learn2reg},
48 pairs).
Since LUMIR is an unsupervised dataset, we use SynthSeg to generate automated segmentation masks delineating 35 cortical and subcortical structures.
We report mean Dice ($\uparrow$) for segmentation benchmarks and mean
Target Registration Error in mm ($\downarrow$) for NLST.

\subsection{Main Results}
\label{sec:main_results}

Table~\ref{tab:main} summarizes the comparison.
On LUMIR, \lm\ outperforms Adam ($0.8715$ vs.\ $0.8650$), with performance per pair improving on $36/38$ pairs, including a higher worst-case Dice ($0.8304$ vs.\ $0.8066$), indicating improved robustness.
OASIS shows a similar trend: \lm\ achieves $0.8293$ against Adam's $0.8028$, a gap of $+0.027$.
On Abdomen-L2R, \lm\ essentially matches Adam ($0.7615$ vs.\ $0.7609$) across all four organ labels (Table~\ref{tab:abdomen_perlabel}).
NLST is the exception: Adam ($0.85$ mm TRE) remains ahead, with \lm\ reaching $1.68$ mm.
However, we note that the average TRE is inflated by a single outlier pair, and excluding it, \lm\ achieves $0.95$ mm versus Adam's $0.83$ mm.
The performance improvement of \lm\ is significantly pronounced for brain datasets, but does not degrade performance on other organ systems.

\begin{table}[t]
  \centering
  \caption{Optimizer comparison across four registration benchmarks.
           Mean Dice ($\uparrow$) for LUMIR, OASIS, Abdomen-L2R;
           mean TRE in mm ($\downarrow$) for NLST.
           ``Pairs won'' counts image pairs where the method beats Adam.
           Best result per dataset in \textbf{bold}.}
  \label{tab:main}
  \vskip 0.1in
  \small
  \begin{tabular}{lccc}
    \toprule
    \textbf{Dataset} & \textbf{Method} & \textbf{Mean} & \textbf{Std} \\
    \midrule
    \multirow{2}{*}{LUMIR (Dice $\uparrow$)}
      & Adam                  & 0.8650 & 0.0201  \\
      & {Levenberg-Marquardt}     & \textbf{0.8715} & \textbf{0.0163} \\
    \midrule
    \multirow{2}{*}{OASIS (Dice $\uparrow$)}
      & Adam                  & 0.8028 & 0.0333  \\
      & {Levenberg-Marquardt}     & \textbf{0.8293} & \textbf{0.0313} \\
    \midrule
    \multirow{2}{*}{NLST (TRE $\downarrow$)}
      & Adam         & \textbf{0.85}  & \textbf{0.28}  \\
      & {Levenberg-Marquardt}            & 1.68  & 2.16  \\
    \midrule
    \multirow{2}{*}{Abdomen-L2R (Dice $\uparrow$)}
      & Adam                  & 0.7609 & \textbf{0.1495}  \\
      & {Levenberg-Marquardt}     & \textbf{0.7615} & {0.1516} \\
    \bottomrule
  \end{tabular}
\end{table}

\begin{table}[t]
  \centering
  \caption{Per-label Dice on Abdomen-L2R (4 organs, 48 pairs).
           \lm\ performs similarly to Adam across all labels.}
  \label{tab:abdomen_perlabel}
  \vskip 0.1in
  \small
  \begin{tabular}{lcc}
    \toprule
    \textbf{Label} & \textbf{Adam} & \textbf{LM} \\
    \midrule
    Spleen       & 0.8344 & 0.8325 \\
    Right kidney & 0.7288 & 0.7211 \\
    Left kidney  & 0.7594 & 0.7662 \\
    Liver        & 0.7208 & 0.7262 \\
    \bottomrule
  \end{tabular}
\end{table}

\subsection{The Damping Cliff: Sensitivity Analysis of \lm\ Hyperparameters}
\label{sec:sensitivity}

We select the default \lm\ hyperparameters ($\lambdainit=0.006$,
$\lambdainc=1.5$, $\lambdadec=0.975$) via a Bayesian optimization search
(Optuna, 50 trials) on LUMIR, starting from these initial values.
To confirm generality, we repeated independent Optuna searches on all four
benchmarks (OASIS, NLST, Abdomen-L2R); in every case the search converges
to nearly identical configurations ($\lambdainc \in [1.5, 2.0]$,
$\lambdadec \approx 0.97$--$0.98$), confirming that these are properties
of the optimizer--task interaction rather than dataset-specific artifacts
(see also \cref{sec:transfer}).

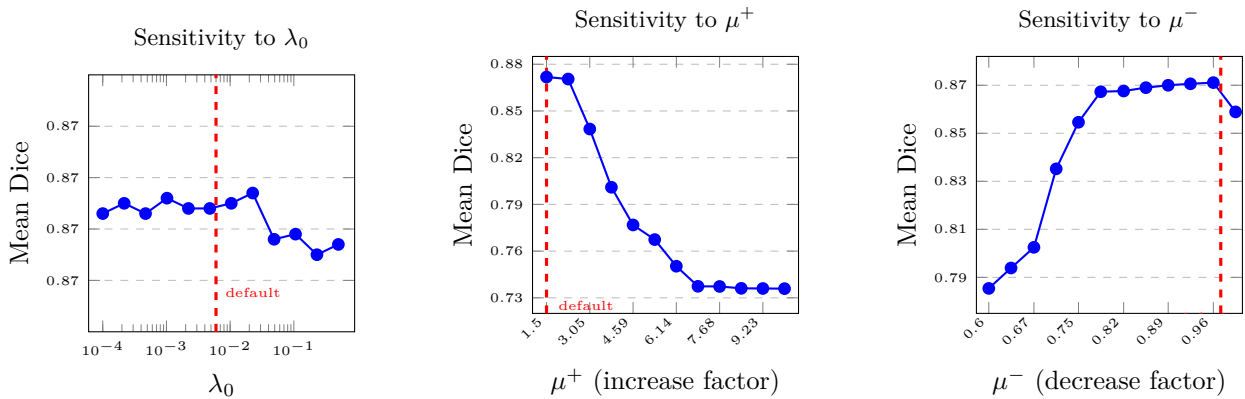
\begin{figure}[t]
\centering
% --- lambda_init (log scale) ---
\begin{tikzpicture}
\begin{axis}[
    width=0.31\textwidth, height=5.0cm,
    xmode=log,
    xlabel={$\lambdainit$},
    ylabel={Mean Dice},
    xmin=0.00006, xmax=0.9,
    ymin=0.869, ymax=0.874,
    xtick={0.0001, 0.001, 0.01, 0.1},
    xticklabels={$10^{-4}$, $10^{-3}$, $10^{-2}$, $10^{-1}$},
    xticklabel style={font=\tiny},
    ytick={0.870, 0.871, 0.872, 0.873},
    yticklabel style={font=\tiny},
    ymajorgrids=true, grid style=dashed,
    title={Sensitivity to $\lambdainit$},
    title style={font=\small}
]
\addplot[color=blue, mark=*, thick] coordinates {
    (0.0001,    0.8713)
    (0.000217,  0.8715)
    (0.000470,  0.8713)
    (0.001020,  0.8716)
    (0.002214,  0.8714)
    (0.004801,  0.8714)
    (0.010414,  0.8715)
    (0.022589,  0.8717)
    (0.048996,  0.8708)
    (0.106275,  0.8709)
    (0.230515,  0.8705)
    (0.500000,  0.8707)
};
\addplot[color=red, dashed, very thick] coordinates {(0.006, 0.869) (0.006, 0.874)};
\node[red, font=\tiny, anchor=south west] at (axis cs: 0.006, 0.8695) {default};
\end{axis}
\end{tikzpicture}%
\hfill
% --- lambda_increase_factor ---
\begin{tikzpicture}
\begin{axis}[
    width=0.31\textwidth, height=5.0cm,
    xlabel={$\lambdainc$ (increase factor)},
    ylabel={Mean Dice},
    xmin=1.0, xmax=10.5,
    ymin=0.72, ymax=0.885,
    xtick={1.5, 3.045, 4.591, 6.136, 7.682, 9.227},
    xticklabel style={font=\tiny, rotate=45, anchor=east},
    ytick={0.73,0.76,0.79,0.82,0.85,0.88},
    yticklabel style={font=\tiny},
    ymajorgrids=true, grid style=dashed,
    title={Sensitivity to $\lambdainc$},
    title style={font=\small}
]
\addplot[color=blue, mark=*, thick] coordinates {
    (1.500,  0.8718)
    (2.273,  0.8705)
    (3.045,  0.8384)
    (3.818,  0.8010)
    (4.591,  0.7768)
    (5.364,  0.7674)
    (6.136,  0.7503)
    (6.909,  0.7374)
    (7.682,  0.7373)
    (8.455,  0.7361)
    (9.227,  0.7360)
    (10.000, 0.7359)
};
\addplot[color=red, dashed, very thick] coordinates {(1.5, 0.72) (1.5, 0.885)};
\node[red, font=\tiny, anchor=north west] at (axis cs: 1.6, 0.734) {default};
\end{axis}
\end{tikzpicture}%
\hfill
% --- lambda_decrease_factor ---
\begin{tikzpicture}
\begin{axis}[
    width=0.31\textwidth, height=5.0cm,
    xlabel={$\lambdadec$ (decrease factor)},
    ylabel={Mean Dice},
    xmin=0.58, xmax=1.01,
    ymin=0.775, ymax=0.882,
    xtick={0.6, 0.673, 0.745, 0.818, 0.890, 0.963},
    xticklabel style={font=\tiny, rotate=45, anchor=east},
    ytick={0.79,0.81,0.83,0.85,0.87},
    yticklabel style={font=\tiny},
    ymajorgrids=true, grid style=dashed,
    title={Sensitivity to $\lambdadec$},
    title style={font=\small}
]
\addplot[color=blue, mark=*, thick] coordinates {
    (0.600, 0.7854)
    (0.636, 0.7939)
    (0.673, 0.8025)
    (0.709, 0.8352)
    (0.745, 0.8546)
    (0.781, 0.8673)
    (0.818, 0.8676)
    (0.854, 0.8690)
    (0.890, 0.8700)
    (0.926, 0.8706)
    (0.963, 0.8711)
    (0.999, 0.8589)
};
\addplot[color=red, dashed, very thick] coordinates {(0.975, 0.775) (0.975, 0.882)};
\node[red, font=\tiny, anchor=north] at (axis cs: 0.952, 0.779) {default};
\end{axis}
\end{tikzpicture}
\caption{Sensitivity of mean Dice on LUMIR (38 pairs) to all three \lm\
         hyperparameters. Our defaults ($\lambdainit=0.006$,
         $\lambdainc=1.5$, $\lambdadec=0.975$, red dashed) were selected
         via Bayesian optimization on LUMIR and are marked in each panel.
         \textbf{Left}: $\lambdainit$ is entirely insensitive---Dice varies
         by $<0.001$ over a $5000\times$ range.
         \textbf{Center}: $\lambdainc$ has a catastrophic cliff between $2.3$
         and $3.0$; our default of $1.5$ sits at the safe peak.
         \textbf{Right}: $\lambdadec$ shows monotone improvement up to
         $0.963$ then a small drop; our default of $0.975$ sits in the
         optimal plateau.}
\label{fig:sensitivity}
\end{figure}

Figure~\ref{fig:sensitivity} shows a sweep of all three hyperparameters on LUMIR,
with the remaining parameters fixed to their defaults.
\paragraph{$\lambdainit$ is entirely insensitive.}
Dice varies by less than $0.001$ across a $5000\times$ range
($\lambdainit \in [10^{-4},\, 0.5]$), because the adaptive schedule in
Eq.~\eqref{eq:lambda-update} rapidly moves $\lambda$ away from any initial
value within the first few iterations.
The default of $0.006$ is representative, but any value in this range is
equally valid.
\paragraph{A catastrophic cliff in $\lambdainc$.}
Dice stays above $0.870$ for $\lambdainc \leq 2.27$, then collapses
by $-0.033$ between $2.3$ and $3.0$, and saturates near $0.736$ above $3.8$; no better
than identity.
Our default of $\lambdainc = 1.5$ lies at the safe peak of this curve,
yielding the best observed Dice of $0.872$.
The mechanism is self-amplifying: a large increase factor aggressively
overdamps on any bad step, driving $\lambda$ upward until every subsequent
update is near-zero.
The safe operating range spans only $\lambdainc \in [1.5, 2.3]$ across all datasets.
\paragraph{Moderate, monotone sensitivity of $\lambdadec$.}
Dice increases monotonically from $0.785$ at $\lambdadec=0.6$ to $0.871$
at $\lambdadec=0.963$, then drops slightly at $\lambdadec=0.999$.
Our default of $0.975$ sits near the top of this plateau, yielding $0.871$.
Too small a decrease factor drops $\lambda$ too quickly after good steps,
overcommitting to Newton-like steps before the curvature estimate is reliable.

\begin{figure}[h]
  \centering
  \includegraphics[width=0.42\textwidth]{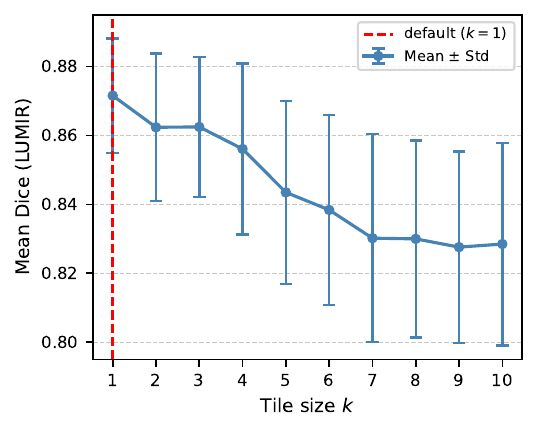}
  \caption{Mean Dice ($\pm$ std) on LUMIR (38 pairs) as tile size $k$ varies
           from 1 to 10, with all other \lm\ hyperparameters fixed to their
           defaults. Pointwise updates ($k{=}1$, red dashed) achieve the best
           mean Dice ($0.8716$) and lowest variance; performance drops
           $-0.009$ at $k{=}2$ and degrades monotonically to $0.828$ by
           $k{=}7$--$10$, with standard deviation rising throughout.}
  \label{fig:tilesize}
\end{figure}

\paragraph{Tile size.}
The \lm\ Hessian approximation can pool over spatial tiles of side $k$,
forming a $d\times d$ local covariance matrix instead of the rank-1
per-voxel outer product.
Figure~\ref{fig:tilesize} sweeps $k=1\ldots10$ on all 38 LUMIR pairs with
all other hyperparameters fixed to their defaults.
Pointwise updates ($k=1$) achieve the best mean Dice ($0.8716$) and lowest
standard deviation ($0.017$); performance drops by $-0.009$ at $k=2$
and degrades roughly monotonically to $0.828$ at $k=7$--$10$, with
standard deviation rising throughout.
Aggregating over a neighbourhood blurs the per-voxel curvature signal,
producing a spatially inconsistent step size that hurts convergence;
we therefore fix $k=1$ for all experiments.

\begin{table}[h]
  \centering
  \caption{Cross-dataset transfer of LUMIR-tuned \lm\ hyperparameters
           compared to per-dataset Optuna search.
           ``LM-transfer'' applies LUMIR parameters directly with no additional
           tuning; ``LM-tuned'' uses dataset-specific search.}
  \label{tab:transfer}
  \vskip 0.1in
  \small
  \begin{tabular}{lccc}
    \toprule
    \textbf{Dataset} & \textbf{Adam} & \textbf{LM-transfer} & \textbf{LM-tuned} \\
    \midrule
    LUMIR (Dice)    & 0.8650 & ---  & 0.8715 \\
    OASIS (Dice)    & 0.8028 & 0.8289 & 0.8293 \\
    NLST (TRE, mm)  & 0.85   & 1.68$^\ddagger$ & 1.68 \\
    Abdomen (Dice)  & 0.7609 & 0.7615 & 0.7615 \\
    \bottomrule
    \multicolumn{4}{l}{\footnotesize $\ddagger$ Inflated by one outlier pair; excluding it: $0.95$ mm vs.\ $0.83$ mm for Adam.}
  \end{tabular}
\end{table}

\subsection{Hyperparameter Transfer and Universality}
\label{sec:transfer}

A practical concern for any optimizer is whether its hyperparameters must
be re-tuned per dataset.
They do not: the optimal \lm\ configuration from LUMIR
($\lambdainit=0.006$, $\lambdainc=1.5$, $\lambdadec=0.975$,
\texttt{tile\_size}=1) transfers directly to OASIS, NLST, and Abdomen-L2R
without modification (Table~\ref{tab:transfer}).
Independent Optuna searches on each target dataset converge to nearly
identical configurations ($\lambdainc \in [1.5, 2.0]$,
$\lambdadec \approx 0.97$--$0.98$), confirming that these are properties
of the optimizer--task interaction rather than dataset-specific artifacts.

\subsection{Step Rejection and Damping Cap}
\label{sec:rejection}

The adaptive damping rule (\cref{eq:lambda-update}) increases $\lambda$
after any bad step, but does not \emph{undo} the step itself.
A natural extension is to additionally restore the warp and retry with increased damping when
the loss degrades beyond a threshold.
Concretely, a step is rejected if
\begin{equation}
  \mathcal{L}^{(t)} - \mathcal{L}^{(t-1)} > \tau \cdot
  \bigl|\mathcal{L}^{(t-1)} - \mathcal{L}^{(t-2)}\bigr|,
  \label{eq:rejection}
\end{equation}
with tolerance ratio $\tau$.
On rejection the warp is restored, $\lambda$ is multiplied by $\lambdainc$,
and the step is retried up to 10 times.
A critical safeguard is a cap $\lambda_{\max}$: without it, repeated rejections
drive $\lambda$ to infinity, collapsing every subsequent update to near-zero.

We ablate two settings ($\tau=1.0$, cap $=1.0$ vs.\ no cap) across all four
datasets using the default \lm\ config.
Table~\ref{tab:rejection} reports the results.
\begin{table}[t]
  \centering
  \caption{Effect of step rejection on default \lm.
           All conditions use $\tau=1.0$.
           Without a cap, rejection is consistently harmful across all datasets.
           With $\lambda_{\max}=1.0$, rejection is dataset-dependent:
           neutral on LUMIR and OASIS, mildly harmful on Abdomen-L2R (48 pairs),
           and beneficial on NLST by suppressing one catastrophic outlier.
           Best result per dataset in \textbf{bold}.}
  \label{tab:rejection}
  \vskip 0.1in
  \small
  \begin{tabular}{llcc}
    \toprule
    \textbf{Dataset} & \textbf{Condition} & \textbf{Mean} & \textbf{Std} \\
    \midrule
    \multirow{3}{*}{LUMIR (Dice $\uparrow$)}
      & \textbf{No rejection}          & \textbf{0.8715} & \textbf{0.0163} \\
      & + rejection, cap $= \infty$    & 0.8654          & 0.0303 \\
      & + rejection, cap $= 1.0$       & 0.8709          & 0.0170 \\
    \midrule
    \multirow{3}{*}{OASIS (Dice $\uparrow$)}
      & \textbf{No rejection}          & \textbf{0.8028} & 0.0333 \\
      & + rejection, cap $= \infty$    & 0.7362          & 0.0899 \\
      & + rejection, cap $= 1.0$       & 0.8014          & 0.0343 \\
    \midrule
    \multirow{2}{*}{Abdomen-L2R (Dice $\uparrow$)}
      & \textbf{No rejection}          & \textbf{0.7615} & 0.1516 \\
      & + rejection, cap $= 1.0$       & 0.7508          & 0.1529 \\
    \midrule
    \multirow{2}{*}{NLST (TRE mm $\downarrow$)}
      & No rejection                   & 1.679           & 2.165  \\
      & \textbf{+ rejection, cap $= 1.0$} & \textbf{1.632} & \textbf{0.593} \\
    \bottomrule
  \end{tabular}
\end{table}
The uncapped case is consistently harmful across all four datasets,
mirroring the runaway $\lambda$ phenomenon seen with large $\lambdainc$.
On LUMIR, removing the cap costs $-0.006$ Dice and nearly doubles standard deviation
(std $0.030$ vs.\ $0.016$); on OASIS the drop is $-0.067$ with standard deviation
nearly tripling (std $0.090$ vs.\ $0.033$).
Capping at $\lambda_{\max}=1.0$ eliminates this failure mode in all cases.

With the cap, outcomes are dataset-dependent.
On LUMIR and OASIS, rejection is effectively neutral ($-0.0006$ and
$-0.0014$ Dice, respectively).
On Abdomen-L2R (48 cross-modal pairs), rejection is mildly harmful
($-0.011$ Dice; winning only 11/48 pairs).
We find that the rejection mechanism is overly conservative on these harder cross-modal pairs, repeatedly rejecting useful steps and slowing convergence.

NLST presents the most interesting case.
Rejection with $\lambda_{\max}=1.0$ reduces mean TRE from $1.68$ to $1.63$\,mm,
but the primary driver is outlier suppression: one pair drops from $8.16$ to
$2.75$\,mm while $9$ of $10$ pairs slightly regress ($+0.11$--$1.63$\,mm).
The rejection mechanism trades a small average regression for a dramatic
worst-case improvement, reflecting that the outlier pair had a genuinely
ill-conditioned step that rejection correctly identifies and dampens.

\textbf{Recommendation.} The cap $\lambda_{\max}$ is essential and prevents
catastrophic divergence in all cases.
Whether to enable rejection ($\tau=1.0$, $\lambda_{\max}=1.0$) is
dataset-dependent: it is beneficial when outlier robustness is the priority
(e.g.\ NLST), and neutral or mildly harmful otherwise.
We disable it by default to provide good asymptotic convergence on average and have enabled robust defaults for the rejection criterion if worst-case cost minimization is desired.

\subsection{Memory and Runtime}
\label{sec:efficiency}

Table~\ref{tab:bench} confirms the theoretical memory advantage.
Because \lm\ carries no moment buffers, it saves $11$--$25\%$ peak GPU
memory depending on volume size, and is never slower than Adam.
The speedup is most pronounced at small volumes ($1.67\times$ at $N=64$,
where Adam's moment-update overhead dominates) and narrows to $\sim 4\%$
at larger $N$ where both methods are compute-bound.
At $N=512$, \lm\ saves $\approx 3{,}024$\,MB.

\begin{table}[t]
  \centering
  \caption{Peak GPU memory (MB) and time per step (s) on an NVIDIA A6000,
           for cubic volumes of side $N$.
           \lm\ is never slower and consistently uses less memory.
           Best result in \textbf{bold}.}
  \label{tab:bench}
  \vskip 0.1in
  \small
  \begin{tabular}{rcccccc}
    \toprule
    & \multicolumn{2}{c}{\textbf{Peak memory (MB)}} & & \multicolumn{2}{c}{\textbf{Time/step (s)}} \\
    \cmidrule{2-3} \cmidrule{5-6}
    $N$ & Adam & \textbf{LM} & Mem.\ saved & Adam & \textbf{LM} & Speedup \\
    \midrule
     64 &   61.1 & \textbf{55.9} &  8.6\% & 0.593 & \textbf{0.356} & $1.67\times$ \\
    128 &  433.1 & \textbf{384.9} & 11.1\% & 0.605 & \textbf{0.606} & $1.00\times$ \\
    192 & 1444.1 & \textbf{1282.7} & 11.2\% & 1.085 & \textbf{1.062} & $1.02\times$ \\
    256 & 3400.1 & \textbf{3022.1} & 11.1\% & 2.127 & \textbf{2.061} & $1.03\times$ \\
    320 & 6638.1 & \textbf{5898.0} & 11.1\% & 3.937 & \textbf{3.801} & $1.04\times$ \\
    384 & 5192.1 & \textbf{3916.4} & 24.6\% & 6.161 & \textbf{5.936} & $1.04\times$ \\
    448 & 8246.1 & \textbf{6218.3} & 24.6\% & 9.730 & \textbf{9.328} & $1.04\times$ \\
    512 & 12296.1 & \textbf{9272.1} & 24.6\% & 14.484 & \textbf{13.952} & $1.04\times$ \\
    \bottomrule
  \end{tabular}
\end{table}

% ============================================================
\section{Discussion}
\label{sec:discussion}
% ============================================================
We discuss a few implications of our experiments and findings.

\paragraph{When does \lm\ help?}
The per-voxel Hessian approximation is most reliable for \textit{in-vivo} brain MRI registration, where the structures of interest have reasonable contrast and have a huge variability in shape (i.e. laminar and curvilinear structures like the cortex, white matter, and sulcal CSF, and amoeboid structures like the ventricles, caudate nucleus, and putamen).
In noisier and sparser settings (cross-modal Abdomen-L2R, highly deformed lung CT), Adam's momentum provides robustness that the \lm\ estimate cannot match on the hardest pairs.

\paragraph{Rehabilitating second-order methods.}
The most striking finding is not that \lm\ is better in aggregate, but that
the optimizer harbors a sharp, self-amplifying failure mode in its damping
schedule.
Dice remains above $0.870$ for $\lambdainc \leq 2.27$, then collapses by
$-0.033$ between $2.3$ and $3.0$, eventually saturating near $0.736$ which is only slightly
better than the identity warp.
A large increase factor aggressively overdamps on any bad step, driving
$\lambda$ upward until every subsequent update is near-zero, locking the warp
in place.
The safe operating range is narrow: $\lambdainc \in [1.5, 2.3]$ across all
four datasets, and an uncapped step-rejection mechanism produces the same
runaway behaviour.

\paragraph{Recommended configuration.}
Based on our sensitivity analysis and cross-dataset transfer experiments,
we recommend the following universal \lm\ settings for diffeomorphic
registration:
\begin{center}
\small
\begin{tabular}{ll}
  \toprule
  $\lambdainit$ & $0.01$ \quad (insensitive; any value in $[10^{-4}, 0.5]$ works) \\
  $\lambdainc$  & $1.5$ \quad (\emph{critical}: must stay $< 2.3$) \\
  $\lambdadec$  & $0.975$ \quad (keep in $[0.78, 0.97]$) \\
  \texttt{tile\_size} & $1$ \quad (pointwise; lower cost, no accuracy loss) \\
  \bottomrule
\end{tabular}
\end{center}
These parameter settings are implemented as the default configuration in the FireANTs implementation.

% % ============================================================
% \section{Related Work}
% \label{sec:related}
% % ============================================================

% Classical diffeomorphic registration methods including
% ANTs~\citep{avants2008ants}, SyN~\citep{avants2008syn}, and
% Demons~\citep{vercauteren2009diffeomorphic} rely on first-order test-time
% optimization.
% Learning-based methods such as VoxelMorph~\citep{balakrishnan2019voxelmorph}
% and LapIRN~\citep{mok2020fast} use second-order information as closed-form
% amortized solvers during training, not as test-time optimizers.
% FireANTs~\citep{jena2025fireants} is the direct predecessor of this work;
% we extend it with a second-order optimizer that exploits the local structure
% of the registration gradient.
% Second-order methods were studied in classical registration
% contexts~\citep{christensen1996deformable}, but their systematic comparison
% against modern first-order optimizers on GPU-based test-time diffeomorphic
% registration is, to our knowledge, new.
% Complementary directions include learned optimizers~\citep{andrychowicz2016learning}
% and curvature-aware preconditioning~\citep{mescheder2017projective}.

% ============================================================
\section{Conclusion}
\label{sec:conclusion}
% ============================================================

We introduced a factored Levenberg--Marquardt optimizer for diffeomorphic
image registration in FireANTs, built on a per-voxel rank-1 Hessian
approximation that makes second-order updates computationally competitive
with Adam.
The central insight is a \emph{damping cliff}: the increase factor $\lambdainc$
must be kept below $2.3$ or registration collapses.
With a corrected configuration tunable from a single dataset and
transferable across modalities, \lm\ matches or outperforms Adam on three
of four benchmarks while using less memory.
We hope this encourages registration on larger datasets owing to the availability of an efficient, trust-region based adaptive optimizer without sacrificing performance.

\bibliography{references}
\bibliographystyle{tmlr}

\clearpage
\appendix
\section{Proof of the Factored Levenberg--Marquardt Update}
\label{sec:proof}

We derive the closed-form pointwise update in ~\cref{eq:tile1} from the
classical \lm\ linear system in ~\cref{eq:lm-classical}.

\paragraph{Setup.}
As argued in Section~\ref{sec:lm}, for a scalar residual $r(\vect{u})$ the
global Jacobian is the gradient vector
$\vect{J} = \nabla_{\vect{u}} L \in \R^{H \times W \times D \times 3}$.
At each voxel $\vect{x}$, we approximate the Gauss--Newton Hessian
\emph{independently} by the rank-1 outer product of the local gradient:
\begin{equation}
  \hat{\vect{H}}_{\vect{x}} = \vect{J}_{\vect{x}} \vect{J}_{\vect{x}}^\top
  = \vect{g}_{\vect{x}} \vect{g}_{\vect{x}}^\top \in \R^{3 \times 3}.
\end{equation}
Substituting into ~\cref{eq:lm-classical} for the local $3$-vector
$\Delta\vect{u}_{\vect{x}}$ gives
\begin{equation}
  \bigl(\vect{g}_{\vect{x}} \vect{g}_{\vect{x}}^\top + \lambda \vect{I}\bigr)
  \Delta\vect{u}_{\vect{x}} = - r \vect{g}_{\vect{x}},
  \label{eq:lm-local}
\end{equation}
where the right-hand side $-\vect{J}_{\vect{x}}^\top$ reduces to
$-\vect{g}_{\vect{x}}$.

\paragraph{Sherman--Morrison inverse.}
The Sherman--Morrison formula states that for an invertible matrix
$\vect{A} \in \R^{n \times n}$ and vectors $\vect{u}, \vect{v} \in \R^n$
with $1 + \vect{v}^\top \vect{A}^{-1} \vect{u} \neq 0$,
\begin{equation}
  \bigl(\vect{A} + \vect{u}\vect{v}^\top\bigr)^{-1}
  = \vect{A}^{-1}
    - \frac{\vect{A}^{-1}\vect{u}\,\vect{v}^\top \vect{A}^{-1}}
           {1 + \vect{v}^\top \vect{A}^{-1}\vect{u}}.
  \label{eq:sherman-morrison}
\end{equation}
We apply this with $\vect{A} = \lambda\vect{I}$,
$\vect{u} = \vect{g}_{\vect{x}}$, $\vect{v} = \vect{g}_{\vect{x}}$:
\begin{align}
  \bigl(\lambda\vect{I} + \vect{g}_{\vect{x}}\vect{g}_{\vect{x}}^\top\bigr)^{-1}
  &= \frac{1}{\lambda}\vect{I}
     - \frac{
         \dfrac{1}{\lambda}\vect{g}_{\vect{x}}\,\vect{g}_{\vect{x}}^\top
         \dfrac{1}{\lambda}
       }{
         1 + \vect{g}_{\vect{x}}^\top \dfrac{1}{\lambda}\vect{g}_{\vect{x}}
       } \notag \\[6pt]
  &= \frac{1}{\lambda}\vect{I}
     - \frac{\vect{g}_{\vect{x}}\vect{g}_{\vect{x}}^\top}
            {\lambda\!\left(\lambda + \|\vect{g}_{\vect{x}}\|^2\right)}.
  \label{eq:sm-result}
\end{align}

\paragraph{Deriving the update.}
Solving ~\cref{eq:lm-local} gives
$\Delta\vect{u}_{\vect{x}}
  = -\bigl(\vect{g}_{\vect{x}}\vect{g}_{\vect{x}}^\top + \lambda\vect{I}\bigr)^{-1}
    r \vect{g}_{\vect{x}}$.
Substituting ~\cref{eq:sm-result} and using
$\vect{g}_{\vect{x}}^\top\vect{g}_{\vect{x}} = \|\vect{g}_{\vect{x}}\|^2$:
\begin{align}
  \Delta\vect{u}_{\vect{x}}
  &= -\left[
       \frac{1}{\lambda}\vect{I}
       - \frac{\vect{g}_{\vect{x}}\vect{g}_{\vect{x}}^\top}
              {\lambda\!\left(\lambda + \|\vect{g}_{\vect{x}}\|^2\right)}
     \right]\vect{g}_{\vect{x}} r \notag \\[4pt]
  &= -\frac{\vect{g}_{\vect{x}} r}{\lambda}
     + \frac{\vect{g}_{\vect{x}} r\,\|\vect{g}_{\vect{x}}\|^2}
            {\lambda\!\left(\lambda + \|\vect{g}_{\vect{x}}\|^2\right)} \notag \\[4pt]
  &= \frac{\vect{g}_{\vect{x}} r}{\lambda}
     \left[
     -1
       + \frac{\|\vect{g}_{\vect{x}}\|^2}
              {\!\left(\lambda + \|\vect{g}_{\vect{x}}\|^2\right)}
     \right] \notag \\[4pt]
  &= \frac{\vect{g}_{\vect{x}} r}{\lambda} \cdot
     \frac{-\!\left(\lambda + \|\vect{g}_{\vect{x}}\|^2\right)
           + \|\vect{g}_{\vect{x}}\|^2}
          {\!\left(\lambda + \|\vect{g}_{\vect{x}}\|^2\right)} \notag \\[4pt]
  &= -\frac{\vect{g}_{\vect{x}} r}{\|\vect{g}_{\vect{x}}\|^2 + \lambda},
\end{align}
which is ~\cref{eq:tile1}. \hfill$\square$

% \paragraph{Remark.}
% The result has an appealing interpretation: the update direction is identical
% to gradient descent ($-\vect{g}_{\vect{x}}$), but the step length is
% \emph{automatically normalized} by the local curvature
% $\|\vect{g}_{\vect{x}}\|^2$.
% Voxels with large gradients (high curvature) receive smaller steps, while
% voxels with near-zero gradients are protected by the floor $\lambda > 0$.
% No $3\times 3$ matrix is ever formed or stored; the only arithmetic
% required is a squared norm and a scalar division per voxel.

\end{document}